\documentclass[a4paper]{article}
\usepackage{INTERSPEECH2022}
\usepackage{graphicx}
\usepackage{booktabs} 
\usepackage{multirow}
\usepackage[ruled]{algorithm2e}
\usepackage[english]{babel}
\usepackage{blindtext}
\usepackage{subcaption}

\title{Binary Early-Exit Network for Adaptive Inference on Low-Resource Devices}
\name{Aaqib Saeed}
\address{Dept. of Computer Science and Technology \\ University of Cambridge}
\email{as3227@cl.cam.ac.uk}

\begin{document}
\maketitle
\begin{abstract}
Deep neural networks have significantly improved performance on a range of tasks with the increasing demand for computational resources, leaving deployment on low-resource devices (with limited memory and battery power) infeasible. Binary neural networks (BNNs) tackle the issue to an extent with extreme compression and speed-up gains compared to real-valued models. We propose a simple but effective method to accelerate inference through unifying BNNs with an early-exiting strategy. Our approach allows simple instances to exit early based on a decision threshold and utilizes output layers added to different intermediate layers to avoid executing the entire binary model. We extensively evaluate our method on three audio classification tasks and across four BNNs architectures. Our method demonstrates favorable quality-efficiency trade-offs while being controllable with an entropy-based threshold specified by the system user. It also results in better speed-ups (latency less than 6ms) with a single model based on existing BNN architectures without retraining for different efficiency levels. It also provides a straightforward way to estimate sample difficulty and better understanding of uncertainty around certain classes within the dataset. 
\end{abstract}


\section{Introduction}
The rapidly increasing size of deep neural networks, along with advancements in design and training, have made them perform well on a broad spectrum of tasks. The large-scale models improve predictive performance but significantly escalate production costs with slower inference, in particular, severely limiting the adoption of deep models on resource-constrained edge devices like wearables with limited battery and computational power. Binary neural networks (BNNs) have become promising methods for obtaining highly compact and efficient models in deployment on resource-constrained devices due to extreme compression and speed-up gains compared to their real-valued counterparts~\cite{courbariaux2016binarized, rastegari2016xnor}. A complementary technique for accelerating deep models is a dynamic input-dependent prediction generation, which has recently gained notable traction and has become widely known as early-exit architecture~\cite{teerapittayanon2016branchynet, bolukbasi2017adaptive, trapeznikov2013supervised, huang2017multi, scardapane2020should, fan2019reducing, schwartz2020right,laskaridis2021adaptive}. 

The key idea behind early-exit models is that the difficulty of classifying an example, and thus the required model capacity to do so, varies greatly in practice. For instance, a speech classifier is likely to be provided with utterances over a wide range of signal-to-noise ratios. The shallow features from initial layers are mostly sufficient to confidently distinguish easy examples of different classes to a certain extent~\cite{schwartz2020right, scardapane2020should, huang2017multi}. These models are capable of trading-off computation and accuracy dynamically on a per-instance basis to efficiently utilize available resources. An early-exit architecture adapts its computation to the difficulty of an example by placing auxiliary decision layers (the ``exits'') at different depths of its main architecture. At inference, an example passes through each exit in a sequential fashion, and a decision rule (or exit threshold) is typically used to decide whether to use the current exit's prediction or to continue. 

We design an early-exit model to reduce inference cost and further speed-up execution of BNNs based on the aforementioned observation. We focus solely on BNNs due to their greater efficiency and minimal model size that is largely suitable for devices with limited memory~\cite{larq}. We address the problem of early-exiting from the perspective of varying model sizes and sample (or instance) difficulty. Our approach provides a simple yet effective way to assess the computational difficulty of making a prediction for a given input and pair it with an optimal exit at a specific depth within the model to reduce uncertainty. As a result, not all instances have to pass through the entire model. The `easy' to classify examples are matched with earlier exits. In comparison, the `harder' examples are tackled by deeper exits to maintain good performance and thus save computation on a per-example basis.

\begin{figure*}[!t]
\centering
\includegraphics[width=0.8\textwidth]{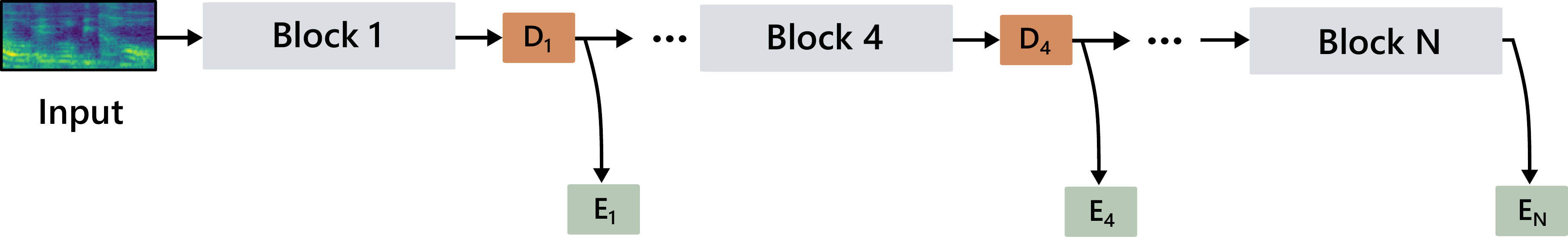}
\caption{Overview of our early-exit binary neural networks. Grey blocks denote a set of binarized convolutional and other layers, orange  and green squares denote decision exits an classification output layers, respectively. Our early-exit model enables adaptive inference for each input sample based on an entropy-based decision threshold to early-exit the input, thus saving computation.}
\label{fig:overview}
\end{figure*}

We apply our method to a broad range of high-performing and efficient binary networks architectures, which we repurpose for different audio recognition tasks. To the best of our knowledge, our work is the first attempt at proposing and studying binary early-exit models for audio recognition. Specifically, we add classification output layers (or also called exits) at varying depths, while maintaining the rest of the architecture as is. We notice that an exit at the $j$th layer is efficient though (seemingly) less effective as compared to later exit $k$, (where $k > j$). Surprisingly, in some cases we achieve similar or even better performance than standard single-exit models, hinting at `over-thinking' phenomena in neural networks in which the predictions based on earlier layers are correct but become incorrect with progressively deeper layers, resulting in wasteful computation~\cite{kaya2019shallow}. We provide high-level illustration of the approach in Figure~\ref{fig:overview}. 

Our approach of unifying binary networks with early exits accelerates inference substantially and results in saving in terms of computation compared to single-exit models while adding very few learnable parameters to the model. Likewise, it does not result in a penalty in terms of training convergence. Our simple strategy is flexible in practice compared to other techniques, for instance, methods that require re-training to achieve an optimal trade-off between speed-up and accuracy. On the contrary, we can train the early-exit BNNs end-to-end and require only the selection of the appropriate value of decision threshold to decide on exiting or further processing of the input. Likewise, the proposed technique is orthogonal to pruning and other strategies for improving neural network efficiency. It can be easily combined to further enhance the computational run-time or energy utilization of the BNNs. 

\section{Method}

We design an early-exit classification architecture with $E$ exits that produces a list of outputs ($p_{e_1}$, $p_{e_2}$, \ldots $p_{e_E}$) where each exit $p_{e_i}$ is a probability distribution over classes. Specifically, for the exits $1 < i <= E$, an exit $e_i$ is potentially more accurate and expensive to compute than the previous exit i.e., $e_{i-1}$. Nevertheless, a slightly less accurate exit can also be used for inference based on a pre-defined threshold of the output to trade-off between accuracy and computation. At inference time, the early-exit model can quantify the exit's confidence on a per-example basis with an entropy of output probability distribution or temperature-scaled softmax score $\mu$. When an instance arrives at an exit, the chosen metric is computed and compared with a pre-defined threshold $S$ to decide to exit earlier or pass on the sample to the next layer. Intuitively, the larger values of the threshold $S$ result in a faster inference but less accurate prediction and vice versa as the earlier exits can be inaccurate to solve the specified learning task. This strategy is also termed as ``early-exiting,'' i.e., for a sample, an output can be produced at any time depending on the computational budget or accuracy. 

We consider binary convolutional networks~\cite{courbariaux2016binarized} as backbone models to augment with early-exit capability for inference on low-resource devices with limited computational power. We represent the weights and input features with $\mathcal{W} \in \mathbb{R}^{k \times k \times c \times o}$ and $\mathcal{X} \in \mathbb{R}^{w \times h \times c}$, with $c$ and $o$ being number of input and output channels, $k$ denoting kernel height and width, while $w$ and $h$ represent spatial dimension of the input. Generally, both the weights and activations of the model are binarized, where convolution operation is performed as $\mathcal{X} * \mathcal{W}$ using bit-wise operations. Previously, various strategies and network architectures have been proposed to tackle degradation in performance due to extreme quantization as compared to their real-valued counterparts. Here, we focus on state-of-the-art binary networks to demonstrate our approach of early-exiting that include QuickNet~\cite{ghosh2017quicknet}, BiRealNet~\cite{liu2018bi}, BinaryDenseNet~\cite{bethge2019back}, and Meliusnet~\cite{bethge2020meliusnet}. The key details of these architectures are summarized in Table~\ref{tab:model_specs}. We utilize Larq Compute Engine for latency benchmarking of binary models. We compute latencies by converting the models to TFLite format using  a device equipped with Snapdragon 855.

Our approach leverages the multilayered structure of models to add output layers after specific intermediate layers. In all the considered architectures, we place five exits in total (including the last) such that early-exit classifiers have adequate capacity. The exits are comprised of a global pooling layer followed by a binary dense layer with hidden units equal to the number of outputs. For model training, the loss function of the $j^{th}$ exit is: $\mathcal{L}_{i}(\mathcal{D}; \theta) = \frac{1}{|\mathcal{D}|} \sum_{(x, y) \in \mathcal{D}} \mathcal{H}(y, f_{j}(x; \theta))$, where, $\mathcal{D}$ is the training set comprising of $(x,y)$ input-label pairs, $\theta$ are the set of learnable model parameters, and $\mathcal{H}$ is the standard cross-entropy loss function. We aggregate the losses from respective output layers to enable earlier layers to act as both useful feature extractors for deeper layers and as good input to their corresponding classifiers. Further, this training regime indicates that every exit layer is trained to perform well on all instances in $\mathcal{D}$. We do not perform early exiting based on decision-threshold (i.e., entropy) during training but employs it for inference. The way our approach works at inference time is shown in Algorithm~\ref{alg:eei}.

\begin{algorithm}[!t]
\caption{Early Exit Inference}
\label{alg:eei}
\scriptsize
\KwIn{Input $x$, Entropy threshold $\delta$}
\KwOut{$y$}
\For{$i \gets 1$ to $E$}
{
    $\hat{y}_{i}$ =  $f_{i}(x; \theta)$\\
    $\epsilon$ = entropy($\hat{y}_{i}$)\\
    \uIf{$\epsilon$ $<$ $\delta$}{
        return $\hat{y}_{i}$
    }
}
return $\hat{y}_{E}$
\end{algorithm}

\renewcommand{\arraystretch}{1.2}

\begin{table}[!htbp]
\centering
\caption{Architectural specification of binary neural networks. We compute latency on a device equipped with Snapdragon 855 with a TFLite model including audio front-end.}
\label{tab:model_specs}
\resizebox{\columnwidth}{!}{%
\begin{tabular}{@{}lccclllll@{}}
\toprule
\multirow{2}{*}{\textbf{Model}} &
  \multirow{2}{*}{\textbf{\#Parameters}} &
  \multicolumn{2}{c}{\textbf{Size (MiB)}} &
  \multicolumn{4}{c}{\textbf{Latency (ms)}} &
   \\ \cmidrule(l){3-9} 
  &
  &
  Binary &
  Float-32 &
  $\mathcal{E}_1$ &
  $\mathcal{E}_2$ &
  $\mathcal{E}_3$ &
  $\mathcal{E}_4$ &
  $\mathcal{E}_5$ \\ \midrule
QuickNet~\cite{ghosh2017quicknet}    & 12.7 M & 2.23 & 48.54 & 2.62 & 2.73 & 2.97 & 3.24 & 3.51 \\

BiRealNet~\cite{liu2018bi}  & 11.2 M & 2.04 & 42.63 & 3.24 & 3.31 & 3.56 & 3.91 & 4.17 \\

BinaryDenseNet~\cite{bethge2019back}  & 4.56 M & 1.85 & 42.63 & 6.71 & 6.14 &  6.19 & 6.24 & 6.31  \\

MeliusNet ~\cite{bethge2020meliusnet} & 6.44 M & 1.94 & 24.56 &  3.24 & 6.58 & 7.08 & 7.37 &  7.59 \\ \bottomrule
\end{tabular}
}
\end{table}

\section{Experiments}
\label{sec:experiments}

\renewcommand{\arraystretch}{1.3}
\begin{table*}[t]
\scriptsize
\begin{subtable}[h]{\textwidth}
\centering
\begin{tabular}{lcccccccccc}
    \hline
    \multirow{2}{*}{\textbf{Model}} &
      \multicolumn{2}{c}{\textbf{Exit 1}} &
      \multicolumn{2}{c}{\textbf{Exit 2}} &
      \multicolumn{2}{c}{\textbf{Exit 3}} &
      \multicolumn{2}{c}{\textbf{Exit 4}} &
      \multicolumn{2}{c}{\textbf{Exit 5}} \\ \cline{2-11} 
     &
      $\mathcal{E}_s$ & $\mathcal{E}_m$ &
      $\mathcal{E}_s$ & $\mathcal{E}_m$ &
      $\mathcal{E}_s$ & $\mathcal{E}_m$ &
      $\mathcal{E}_s$ & $\mathcal{E}_m$ &
      $\mathcal{E}_s$ & $\mathcal{E}_m$  \\ \hline
    QuickNet       & 56.8 & 61.2 & 73.4 & 76.1 & 74.8 & 78.7 & 80.7 & 81.0 & 78.3 & 80.7 \\
    BiRealNet      & 59.1 & 58.0 & 74.6 & 72.9 & 77.2 & 77.7 & 79.7 & 80.4 & 80.2 & 80.3 \\
    BinaryDenseNet & 78.9 & 73.1 & 83.8 & 79.1 & 83.8 & 80.5 & 83.8 & 81.1 & 81.7      & 80.5 \\
    MeliusNet      & 18.0 & 41.2 & 84.3 & 81.7 & 87.9 & 86.3 & 87.5 & 87.4 & 85.9      & 86.3 \\ \hline
\end{tabular}
\caption{MSWC (Micro-EN)}
\end{subtable}
 
\begin{subtable}[h]{\textwidth}
\centering 
\begin{tabular}{lllllllllll}
     \hline
      \multirow{2}{*}{\textbf{Model}} &
      \multicolumn{2}{c}{\textbf{Exit 1}} &
      \multicolumn{2}{c}{\textbf{Exit 2}} &
      \multicolumn{2}{c}{\textbf{Exit 3}} &
      \multicolumn{2}{c}{\textbf{Exit 4}} &
      \multicolumn{2}{c}{\textbf{Exit 5}} \\ \cline{2-11} 
     &
      $\mathcal{E}_s$ & $\mathcal{E}_m$ &
      $\mathcal{E}_s$ & $\mathcal{E}_m$ &
      $\mathcal{E}_s$ & $\mathcal{E}_m$ &
      $\mathcal{E}_s$ & $\mathcal{E}_m$ &
      $\mathcal{E}_s$ & $\mathcal{E}_m$  \\ \hline
    QuickNet       & 65.7 & 86.3 & 92.8 & 93.3 & 95.2 & 94.7 & 94.6 & 95.0 & 94.4 & 95.1 \\
    BiRealNet      & 90.7 & 89.5 & 91.9 & 93.1 & 94.7 & 94.2 & 92.9 & 94.0 & 91.7 & 93.9 \\
    BinaryDenseNet & 65.6 & 93.2 & 93.1 & 95.2 & 94.4 & 95.9 & 91.1 & 96.0 & 93.6 & 95.9 \\
    MeliusNet      & 81.9 & 81.5 & 94.7 & 93.8 & 95.6 & 93.4 & 95.2 & 94.9 & 94.9 & 94.9 \\ \hline
\end{tabular}
\caption{SpeechCommands}
\end{subtable}

\begin{subtable}[h]{\textwidth}
\centering 
\begin{tabular}{lcccccccccc}
      \hline
      \multirow{2}{*}{\textbf{Model}} &
      \multicolumn{2}{c}{\textbf{Exit 1}} &
      \multicolumn{2}{c}{\textbf{Exit 2}} &
      \multicolumn{2}{c}{\textbf{Exit 3}} &
      \multicolumn{2}{c}{\textbf{Exit 4}} &
      \multicolumn{2}{c}{\textbf{Exit 5}} \\ \cline{2-11} 
     &
      $\mathcal{E}_s$ & $\mathcal{E}_m$ &
      $\mathcal{E}_s$ & $\mathcal{E}_m$ &
      $\mathcal{E}_s$ & $\mathcal{E}_m$ &
      $\mathcal{E}_s$ & $\mathcal{E}_m$ &
      $\mathcal{E}_s$ & $\mathcal{E}_m$  \\ \hline
    QuickNet       & 75.7 & 75.8 & 88.8 & 83.0 & 86.2 & 84.3 & 81.3      & 84.9 & 70.9      & 79.0 \\
    BiRealNet      & 72.3 & 70.8 & 82.9 & 73.9 & 87.3 & 80.1 & 81.6      & 76.6 & 76.2 & 78.1 \\
    BinaryDenseNet & 87.6 & 83.5 & 86.3 & 88.6 & 87.8 & 89.9 & 85.2      & 87.3 & 82.2      & 85.0 \\
    MeliusNet      & 70.5 & 74.0 & 89.4 & 87.3 & 91.0 & 90.8 & 77.6 & 88.6 & 84.5      & 88.2 \\ \hline
\end{tabular}
\caption{Voxforge}
\end{subtable}

\caption{Performance (accuracy) analysis of single-exit models $\mathcal{E}_s$ with the corresponding exit in an early-exit model $\mathcal{E}_m$, where all exits are trained in an end-to-end manner. We note that the performance of multi-exit classifiers is inline with when models up to certain intermediate layer are trained independently.}
\label{tab:mswc_table}
\end{table*}

\begin{figure*}[t]
\centering
\includegraphics[width=0.68\textwidth]{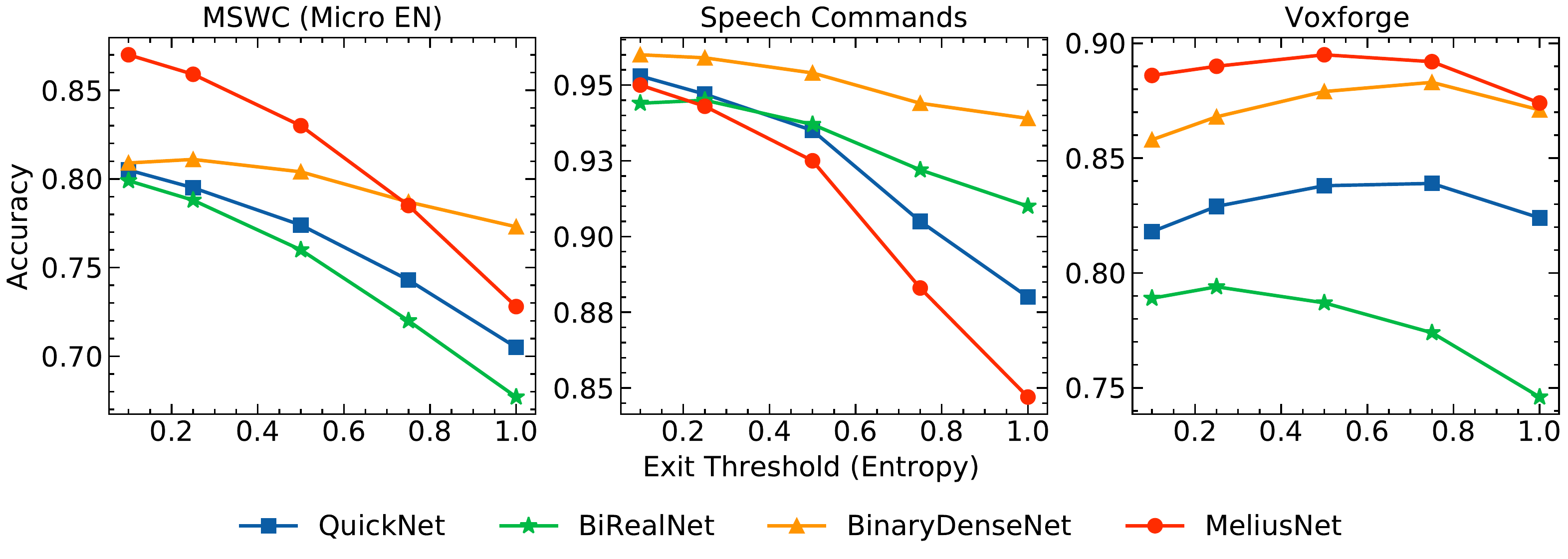}
\caption{Evaluation of early-exit binary neural networks across a range of tasks and architectures for varying entropy thresholds. The early-exit models provide an adaptive exiting strategy for each example and maintain a good balance between exiting early and increasing uncertainty. Our approach also achieves performance close to state-of-the-art floating-point models on Voxforge and SpeechCommands datasets close to $90\%$ and $95\%$ in comparison to $95.4\%$~\cite{shukla2019spoken} and $97.9\%$ ~\cite{seo2021wav2kws}, respectively.}
\label{fig:entropy_accuracy}
\end{figure*}

\subsection{Datasets, Tasks and Implementation Details}
We conduct experiments on various audio recognition tasks ranging from spoken language identification, keyword spotting, to spoken commands recognition. We choose Speech Commands~\cite{warden2018speech} and Voxforge~\cite{maclean2018voxforge}, and MSWC (Micro-EN)~\cite{mazumder2021multilingual} datasets sampled at $16$kHz with each comprising $12$, $6$, and $31$ classes, respectively. In all cases, we utilize the standard train-test splits provided with the corresponding datasets. Given an audio input sequence, we randomly select a one-second of audio segment from an entire audio clip to extract log-compressed Mel-filterbanks with a window size of $25$ ms, a hop size of $10$ ms, and $N = 64$ Mel-spaced frequency bins in the range $60$-$7800$ Hz for $T =  98$ frames, corresponding to $980$ ms. We use these features as input to binary neural networks, i.e., QuickNet, BiRealNet, BinaryDenseNet-28, and MeliusNet. Even though these architectures are originally proposed for vision tasks, the $2$D input structure of Mel-filterbanks enables us to use them without further adjustment. We use the Larq~\cite{larq} framework for implementation of BNNs and utilize Binary Optimizer (Bop)~\cite{helwegen2019latent} and Adam~\cite{kingma2014adam} for training models with a learning rate of $0.001$ and batch size of $128$ for $100$ epochs. During the evaluation, we use each audio sample without any splitting, effectively using a batch size of one to handle input of varying sizes.

\begin{figure*}[!t]
\centering
\subfloat{\includegraphics[width=0.33\textwidth]{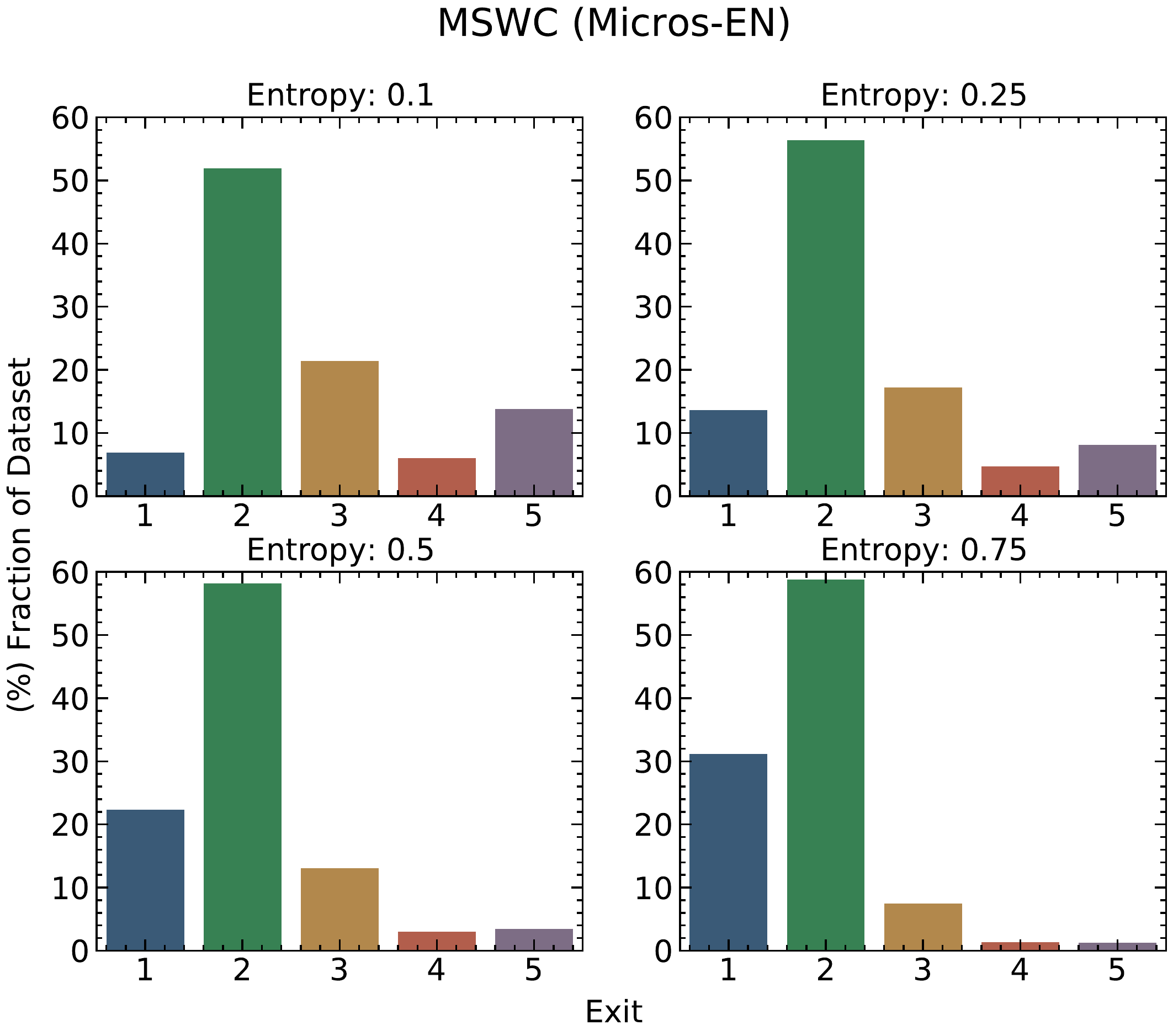}} 
\subfloat{\includegraphics[width=0.33\textwidth]{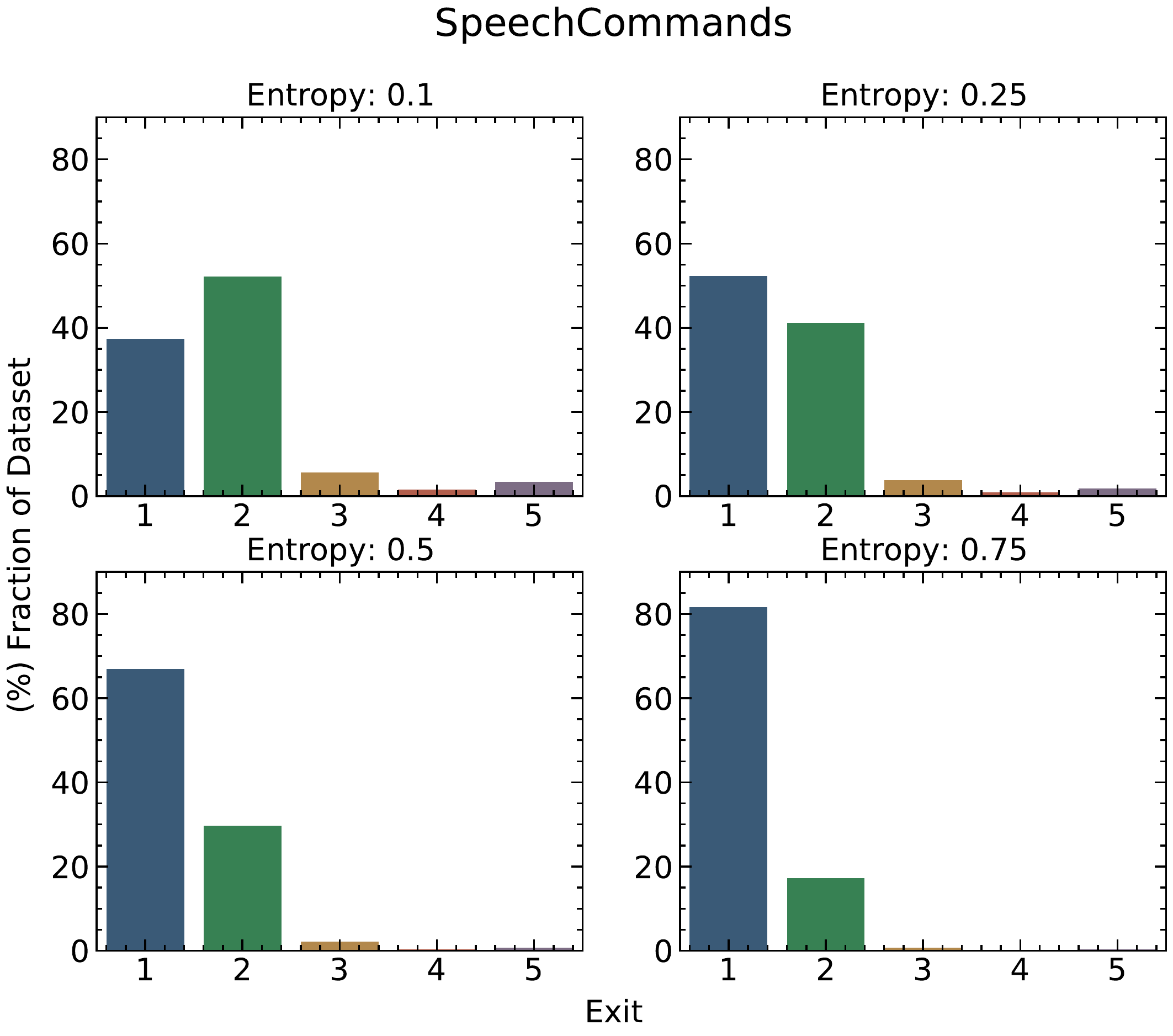}} 
\subfloat{\includegraphics[width=0.33\textwidth]{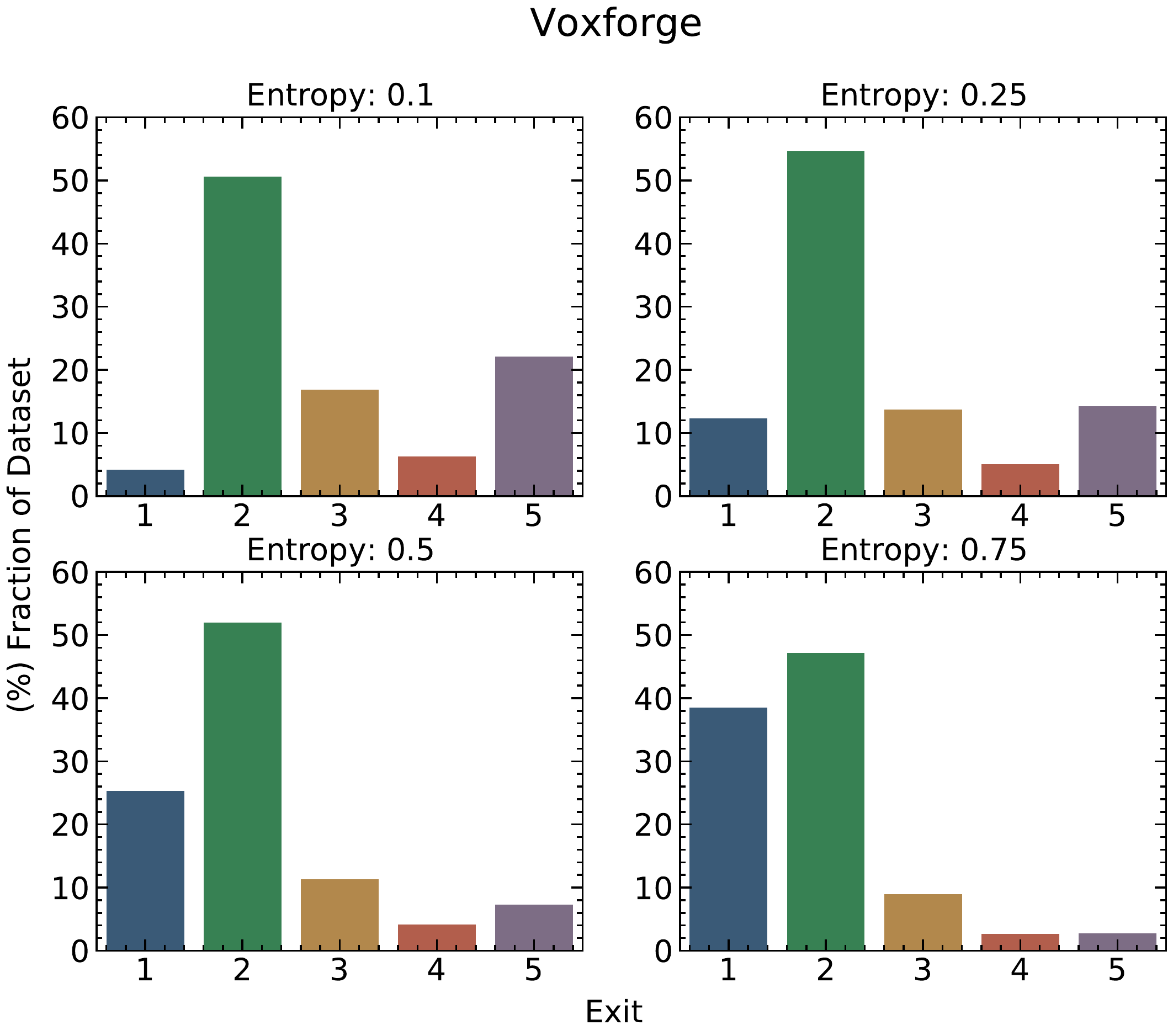}} 
\caption{The number of output samples by exit layer for different entropy thresholds for Meliusnet model. With increasing entropy threshold, a large fraction of samples utilize early exits, where fifth block is the last exit in the model.}
\label{fig:exit_frac}
\end{figure*}

\begin{figure}[t]
\centering
\subfloat{\includegraphics[width=\columnwidth]{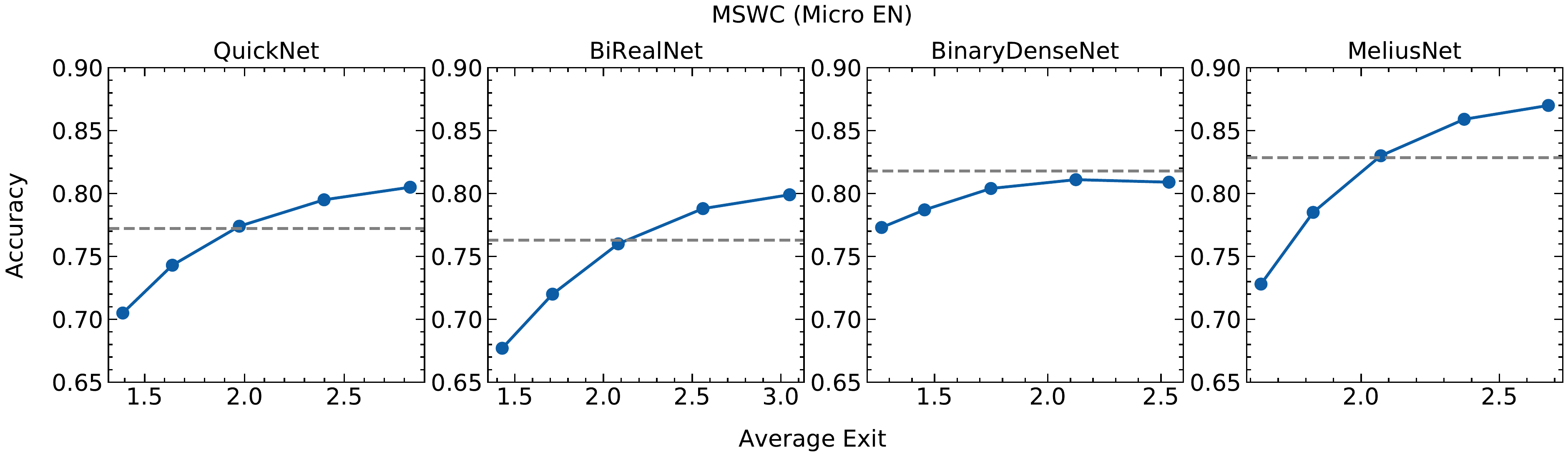}} \\
\subfloat{\includegraphics[width=\columnwidth]{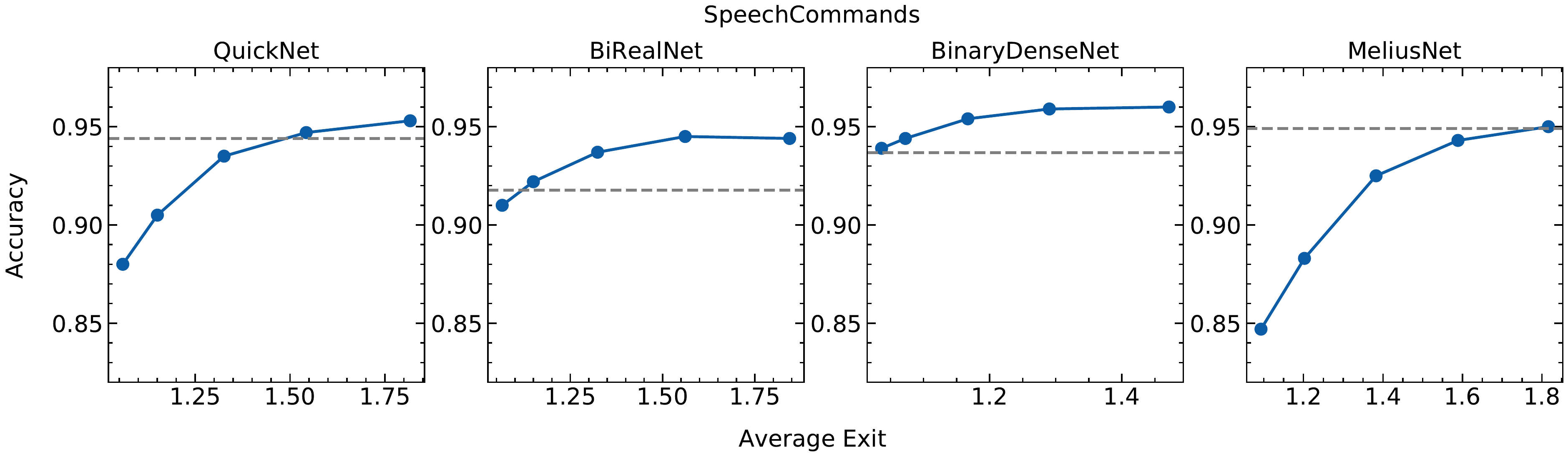}} \\
\subfloat{\includegraphics[width=\columnwidth]{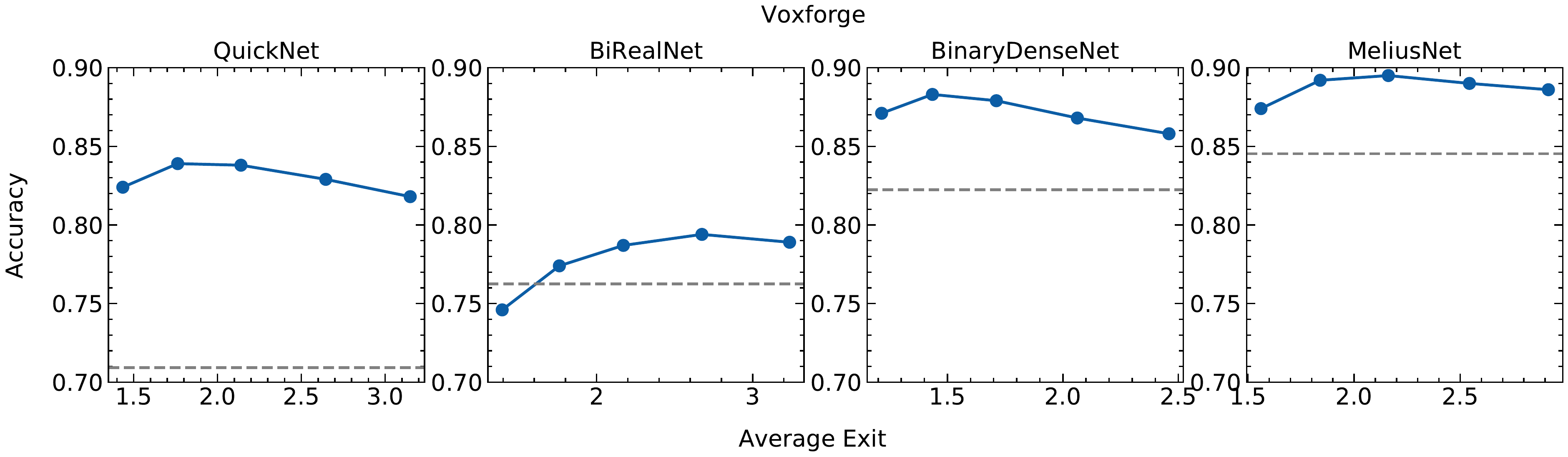}} 
\caption{Average exit and accuracy trade-offs for different binary models. The later exits classify examples with higher accuracy but at an increased computational cost.}
\label{fig:exit_accuracy}
\end{figure}

\subsection{Results}
We start with analyzing the generalization of each exit for considered neural architectures and datasets in Table~\ref{tab:mswc_table}. The exits shown with $\mathcal{E}_s$ are single exit models where a single output layer after some intermediate layer is added while $\mathcal{E}_m$ represents the same exit but trained jointly in an end-to-end fashion. We notice that the performance of individual exits within early-exit models is mostly consistent with the ones trained independently. Further, the single exit model ($\mathcal{E}_s$) can also be seen as efficient baselines, which offer a trade-off between accuracy and speed, but unlike with one model like our approach. 

In Figure~\ref{fig:entropy_accuracy}, we present our key results achieved with different entropy thresholds for each task and architecture. The curves report accuracy achieved while setting a fixed value of entropy $\delta$ with an early-exit model. We use entropy thresholds of $0.1, 0.25, 0.5, 0.75$, and $1.0$. We observe that even with high values of $\delta$, the performance stays considerably stable. In the case of SpeechCommands, for all early-exit BNNs, the accuracy is around $91\%$ when $\delta = 0.5$. Interestingly, on Voxforge, we see a significant difference in performance across architectures. Specifically, the BiRealNet accuracy is lower than the rest and degrades more rapidly with increasing uncertainty. However, MeliusNet performs best with $\delta = 0.5$, achieving around $90\%$ accuracy. On the other hand, we notice that architectures achieve optimal performance with $0.6 < \delta < 0.8$. Both the early and late exits results in lower performance, these results echos the finding of~\cite{kaya2019shallow}, that ``overthinking'' can lead to incorrect predictions.  

\begin{figure}[!ht]
\centering
\includegraphics[width=0.8\columnwidth]{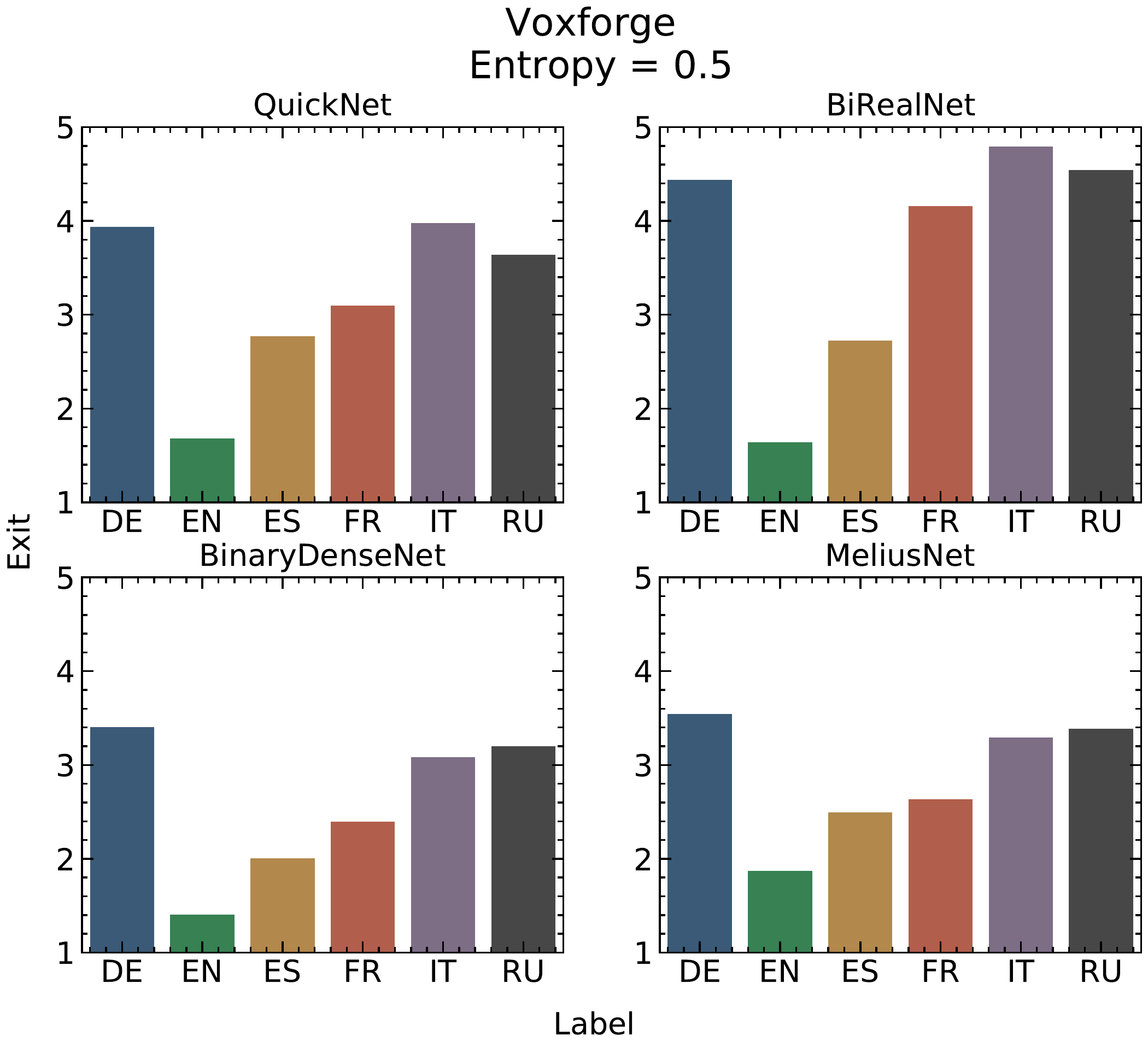}
\caption{Comparison of Voxforge class labels (x-axis) per exit. Each plot shows results with a specific early-exit model for a fixed entropy threshold, i.e., $\delta = 0.5$. }
\label{fig:voxforge_exit_label}
\end{figure}

We show the fraction of instances exiting from each exit of Meliusnet for a fixed entropy threshold in Figure~\ref{fig:exit_frac}. The lower values of $\delta$ force the samples to increasingly utilize later exits. With increasing $\delta$, more examples exit early, thus saving computation. We notice example difficulty varies across tasks. For instance, in the case of SpeechCommands, the first exits generally dominate the inference to classify the majority of examples. However, on MSWC and Voxforge, all exits are mostly utilized. In particular, we observe an interesting pattern for Voxforge that instances mostly either exit early or prefer the last exit with few using the fourth exit. These results highlight that entropy-based exiting optimally selects suitable exit while maintaining a good trade-off between predictive quality and efficiency.    
In Figure~\ref{fig:exit_accuracy}, we show the accuracy of binary models with respect to average exit for various entropy thresholds. The dotted gray line represents the performance of the standard single exit model, where all instances exit from the last layer. We observe consistent results with later exits improving performance at the cost of utilizing more compute, with BinaryDenseNet and MeliusNet achieving higher accuracy as compared to the rest. Furthermore, Figure~\ref{fig:voxforge_exit_label} presents the proportion of instances in the Voxforge test set for each semantic class and models for entropy threshold of $0.5$. We notice that across models, not all classes utilize all exits. In particular, EN is recognized with early exits, while DE and RU largely used later exits. We also observe that in QuickNet and BiRealNet, difficult classes pushed more towards the last layers, on average using the fourth exit.

\section{Conclusions}
We unify binary neural networks with the early-exiting approach for audio recognition on devices with low computational resources. To this end, we present a method to enhance their quality-efficiency trade-offs for efficient inference. Our conceptually simple and architecture-agnostic approach augments standard models with early exits at different intermediate layers to make predictions on easy instances by executing fewer layers. Our experimental evaluation on three different tasks and four neural architectures demonstrates the effectiveness of our approach over standard single exit models while maintaining similar performance at a lower computational budget. Further, our technique provides a simple strategy based on entropy for controlling the speed and accuracy of inference with a single model.  

\bibliographystyle{IEEEtran}
\bibliography{main}
\end{document}